\pdfoutput=1

\documentclass[11pt]{article}

\usepackage[]{acl}

\usepackage{graphicx}
\usepackage{times}
\usepackage{latexsym}
\usepackage{tabularx}  
\usepackage{xcolor}

\definecolor{dkgreen}{rgb}{0,0.6,0}
\definecolor{gray}{rgb}{0.5,0.5,0.5}
\definecolor{mauve}{rgb}{0.58,0,0.82}

\usepackage{listings}
\definecolor{codebackground}{RGB}{240, 240, 235}
\definecolor{LightGray}{gray}{0.9}

\newlength{\msize}
\usepackage[T1]{fontenc}

\usepackage[utf8]{inputenc}
\newcommand{\chris}[1]{{\textcolor{green}{\bf [{\sc chris:} #1]}}}
\newcommand{\demian}[1]{{\textcolor{orange}{\bf [{\sc demian:} #1]}}}

\usepackage{microtype}

\title{News Signals: An NLP Library for Text and Time Series}

\author{Chris Hokamp* \and Demian Gholipour Ghalandari* \and Parsa Ghaffari \\
  Quantexa \\
  \texttt{<firstname><lastname>@quantexa.com}
\\}

\begin{document}
\maketitle

\def\thefootnote{*}\footnotetext{equal contribution}\def\thefootnote{\arabic{footnote}}

\begin{abstract}

We present an open-source Python library for building and using datasets where inputs are clusters of textual data, and outputs are sequences of real values representing one or more time series signals. The \texttt{news-signals} library supports diverse data science and NLP problem settings related to the prediction of time series behaviour using textual data feeds. For example, in the news domain, inputs are document clusters corresponding to daily news articles about a particular entity, and targets are explicitly associated real-valued time series: the volume of news about a particular person or company, or the number of pageviews of specific Wikimedia pages. Despite many industry and research use cases for this class of problem settings, to the best of our knowledge, News Signals is the only open-source library designed specifically to facilitate data science and research settings with natural language inputs and time series targets. In addition to the core codebase for building and interacting with datasets, we also conduct a suite of experiments using several popular Machine Learning libraries, which are used to establish baselines for time series anomaly prediction using textual inputs.

\end{abstract}

\section{Introduction}
\label{sec:introduction}

The natural ordering of many types of data along a time dimension is a consequence of the known physics of our universe. Real-world applications of machine learning often involve data with implicit or explicit temporal ordering. Examples include weather forecasting, market prediction, self-driving cars, and language modeling.

A large body of work on time series forecasting studies models which consume and predict real-valued target signals that are explicitly ordered in time; however, aside from some existing work mainly related to market signal prediction using social media \citep{finnlp-2021-financial,finnlp-2022-financial,arno-etal-2022-next,Li-2014-news-stocks,BingLi-twitter-sa-for-stock-price,kim2016-predicting-cryptocurrency,wang-luo-2021-predicting}, inter alia, the NLP research community has generally not focused on tasks with textual inputs and time series outputs. This is confirmed by the lack of any popular NLP tasks related to time series in result-tracking projects such as nlp-progress\footnote{\url{https://nlpprogress.com/}} and papers-with-code\footnote{\url{https://paperswithcode.com/}}. 

We believe there is potential for novel, impactful research into tasks beyond market signal forecasting, in which textual inputs and real-valued output signals are explicitly organized along a time dimension with fixed-length "ticks". Two reasons for the lack of attention to such tasks to date may be:

\begin{enumerate}
    \item researchers do not have access to canonical NLP datasets for time series forecasting.
    \item data scientists are missing a high level software library for NLP datasets with time series.
\end{enumerate}

\begin{figure}[t]
  \centering
  \includegraphics[width=.95\linewidth]{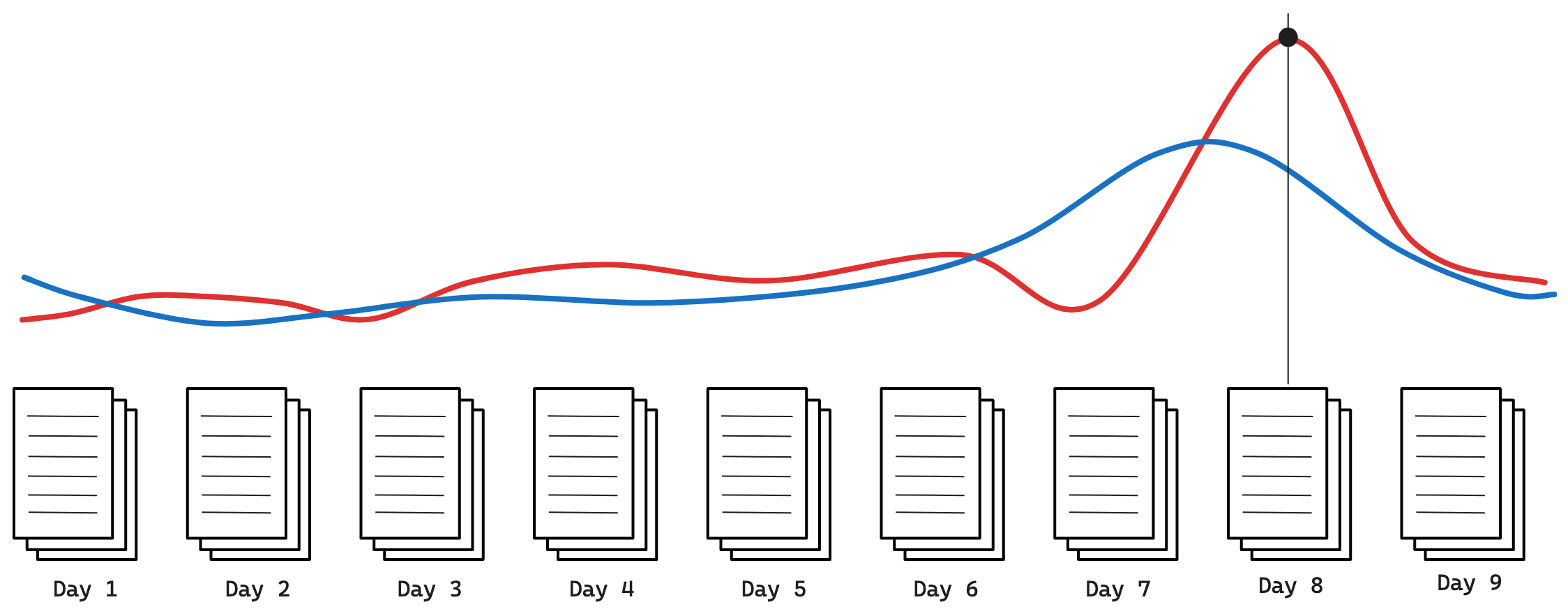}
  \caption{News Signals Datasets: clusters of documents, bucketed by time period, are associated with time series signals. ML models can be trained to predict time series signals using the textual data.}
  \label{fig:news-signals-dataset}
\end{figure}

\noindent Examples of tasks where natural language input can be used to predict a time series signal include:
\begin{itemize}
    \itemsep0em 
    \item weather or pandemic forecasting using social media posts from a recent time period,
    \item market signal prediction using newsfeeds or bespoke textual data feeds,
    \item media monitoring for consumer behavior prediction and forecasting,
    \item forecasting the impact of a news event on the pageviews of a particular website,
\end{itemize}

\noindent and many others. We refer to this general task setting as \texttt{text2signal} (T2S).

\subsection{\texttt{news-signals}}
This work introduces \texttt{news-signals}\footnote{\url{https://github.com/AYLIEN/news-signals-datasets}}, a high-level MIT-licensed software package for building and interacting with datasets where inputs are clusters of texts, and outputs are time series signals (Figure \ref{fig:news-signals-dataset}). Despite the package's news-focused origins, it is built to be a general purpose library for interacting with time-ordered clusters of text and associated time series signals.  

Preparing and utilizing datasets for T2S tasks requires purpose-built software for retrieving and sorting data along the time dimension. In many cases, data will be retrieved from one or more APIs, or web-scraped, further complicating dataset generation pipelines. \texttt{news-signals} exposes an intuitive interface for generating datasets that we believe will be straightforward for any data scientist or developer familiar with the Python data science software stack (see Section \ref{sec:time-indexed-nlp-datasets}).  

\texttt{news-signals} includes tooling for: 
\begin{itemize}
    \itemsep0em 
    \item calling 3rd party APIs to populate signals with text and time series data,
    \item visualizing signals and associated textual data,
    \item extending signals with new time series, feeds, and transformations,
    \item aggregations on textual clusters, such as abstractive and extractive summarization.
\end{itemize}

\texttt{news-signals} provides two primary interfaces: \texttt{Signal} and \texttt{SignalsDataset}. A \texttt{SignalsDataset} represents a collection of related signals. A \texttt{Signal} consists of one or more textual \textbf{feeds}, each connected to one or more time series. Time series have strictly one real value per-tick, while feeds are time-indexed buckets of textual data. For example, a news signal might contain a feed of all articles from a financial source that mention a particular company, linked to multiple time series representing relevant market signals for that company.

\noindent \texttt{news-signals} datasets are designed to be easy to extend with new data sources, entities, and time series signals. In our initial release of the library, we work with three collections of entities: US politicians, NASDAQ-100 companies, and S\&P 500 companies (see section \ref{sec:experiements}). 

The rest of the paper is organized as follows: section \ref{sec:time-indexed-nlp-datasets} gives an overview of library design and Section \ref{sec:signal-apis} describes the \texttt{Signal} and \texttt{SignalsDataset} APIs, the two main interfaces to time-indexed NLP datasets. Section \ref{sec:building-datasets} discusses how datasets can be created. Section \ref{sec:experiements} describes our example datasets, models, and end-to-end experiments, which are open-source, and can be used as templates for new research projects. Section \ref{sec:news-signals-applications} discusses applications, Section \ref{sec:related-work} reviews related work, and Section \ref{sec:conclusion} gives conclusions and directions for the future.

\section{Time-Indexed NLP Datasets}
\label{sec:time-indexed-nlp-datasets}

Traditional NLP and ML datasets consist of iid $ (X, Y) $ pairs. These pairs can be assigned indices, and be operated on by standard pre-processing procedures, such as randomly shuffling and splitting into \texttt{train}, \texttt{dev}, \texttt{test} subsets. However, for time series forecasting and related tasks, inputs are ordered along a time axis, and the distribution of later time steps is typically heavily dependent upon the distribution of earlier time steps; therefore, training, dev and test subsets are usually partitioned and split chronologically to reduce the potential for leakage, introducing additional complexity into data preparation. 

Within the Python data science ecosystem, libraries such as Numpy \citep{harris2020array}, Pandas \citep{mckinney-proc-scipy-2010}, and Pytorch \citep{Paszke_Pytorch_NEURIPS2019} have standardized a syntax for indexing and slicing multi-dimensional matrices and dataframes along axes. When a Pandas dataframe is indexed along a dimension with time-interval semantics, slicing between dates or timestamps is a very useful feature. For example, a user may want to work with the news articles and corresponding time series signals that occurred between particular \texttt{START} and \texttt{END} dates. Pandas in particular includes rich tooling for indexing and slicing datasets along time-indexed axes, and \texttt{news-signals} delegates slice commands and indexing to Pandas, exposing an interface for interacting with datasets using datetime indices\footnote{\url{https://pandas.pydata.org/docs/reference/api/pandas.DatetimeIndex.html}}.

\subsection{\texttt{news-signals} Technical Requirements}

The key technical desiderata we took into consideration when building \texttt{news-signals} are listed below:

\begin{itemize}
    \itemsep0em 
    \item the complexity of data retrieval should be minimized: calling APIs, retrying failed requests, and parsing API output should be invisible to users.
    \item large datasets containing hundreds or thousands of signals, each lasting for thousands of "ticks",  should be straightforward to configure and build.
    \item standard data science libraries such as Pandas should be used as much as possible to reduce maintenance burden over time.
    \item transformations on time series such as anomaly detection or trend/seasonality removal should be straightforward to implement.
    \item the complexity of compressing, saving, and loading datasets locally and remotely should be invisible to users.
    \item new types of signals should be easy to implement.
    \item Signals should be easy to use with standard machine learning libraries.
\end{itemize}

\section{The \texttt{Signal} and \texttt{SignalsDataset} APIs}
\label{sec:signal-apis}

Signals consist of at least one time series coupled with zero or more textual data feeds. Figure \ref{fig:signal-example} shows an example of creating and populating a \texttt{Signal}. Because most functions on the signal class return the signal itself, users can employ a convenient chaining syntax when performing multiple operations on a signal.

\begin{figure}[!tb] 
\begin{lstlisting}
import datetime
from news_signals import signals

# wikidata QID for Twitter
qid = 'Q918'

signal = signals.AylienSignal(
    name='Twitter-Signal',
    params={"entity_ids": [qid]}
)

start = '2023-01-01'
end = '2023-06-01'
# retrieve a timeseries for the count of 
# news articles per-day for this signal
signal = \
    signal(start, end).anomaly_signal()
# sample stories for every day in the signal
signal = signal.sample_stories()

# let's have a look at the biggest anomaly
top_day = signal.anomalies.idxmax()

# what was going on that day?
stories = signal.feeds_df.loc[top_day]['stories']
for s in stories:
    print(s['title'])

# Twitter experiencing outages nationwide
# Twitter experiencing international outages ...
# It's Not Just You, Twitter Is Acting Weird
# : Twitter briefly goes down
# Twitter outage: what happened, ...
#....
\end{lstlisting} 
\caption{Creating and using a news signal\label{fig:signal-example}} 
\end{figure} 

The library retrieves and stores the time series and news stories for the signal, and exposes a Pandas-like API to the underlying dataframes. We can add arbitrary textual data feeds to signals; in figure \ref{fig:signal-example}, \texttt{signal.sample\_stories()} samples stories for every day of the time series (see library documentation on GitHub for more detailed information on how this works).

Once feeds and time series have been initialized, users can perform exploratory data analysis (EDA) in many ways, for example by examining and summarizing the news stories for an anomalous window of the signal's time series, or by plotting the signal.

Signals can also be easily mapped into a single dataframe representation by using the \texttt{.df} property. Signals' dataframe representations contain the textual and time series data associated with a signal, indexed along a \texttt{DatetimeIndex}, but they do not contain metadata such as how the signal is populated from one or more APIs, and transformation semantics such as how anomalies are computed.

Signals automatically differentiate between textual data and time series data types -- for example, when \texttt{signal.plot()} is called, a signal's associated time series are automatically plotted in a multi-line plot. 

\subsection{API integrations}

Most signals require retrieving data from one or more third-party APIs or on-disk datasets. In the current version of \texttt{news-signals}, we provide a deep integration with the Aylien NewsAPI, and additionally implement an interface to the Wikidata pageviews API for building pageview time series for Wikidata items \footnote{\url{https://wikitech.wikimedia.org/wiki/Analytics/AQS/Pageviews}}.

\subsection{The \texttt{SignalsDataset} API}
\label{subsec:signals-datasets}

Individual signals can be grouped into \textit{datasets}. The \texttt{SignalsDataset} is a useful abstraction for working with groups of related signals --- concretely, these might be signals for all politicians from a particular country, or for all companies connected to a specific market subset, such as the NASDAQ-100 or the S\&P-500. Another dataset type could contain signals encapsulating content and time series related to different social media forums, such as Subreddits \citep{wang-luo-2021-predicting}. The number of signals in a dataset can easily number in the hundreds or thousands, so we design a simple configuration DSL using \texttt{yaml} to allow easy construction of large datasets, which is documented in our GitHub repository.

\paragraph{Aylien NewsAPI and Wikimedia APIs}

Because our production use cases for \texttt{news-signals} are focused upon analyzing news data from the Aylien NewsAPI\footnote{Aylien was acquired by Quantexa in February 2023}, the flagship Signal type in \texttt{news-signals} is currently\footnote{as of August 2023} the \texttt{AylienSignal}. This signal type abstracts away API call semantics, allowing users to populate a signal by simply calling \texttt{signal(start\_date, end\_date)}. Of the data sources currently implemented in \texttt{news-signals}, Wikidata is completely free, but the Aylien NewsAPI requires a license key. However, we note that the Aylien NewsAPI currently has a two-week free trial allowing significant free API calls\footnote{https://aylien.com/news-api-signup}, and we hope to implement Signal types for fully public data sources beyond Wikidata in the near future.

\subsubsection{Saving and loading Datasets}

Local and remote serialization and persistence are essential features for dataset-focused libraries, and both Signal and SignalsDataset support saving and loading. We have also implemented persistent on Google Drive and Google Cloud Storage, that only require a remote path to be provided. Datasets are decompressed and cached locally so that the same dataset will not be re-downloaded if it is already available locally.

\paragraph{Library Documentation} Section \ref{sec:signal-apis} has given only a small sample of the \texttt{news-signals} library capabilities, and we refer interested readers to the library documentation on GitHub, which also includes end-to-end example notebooks and video walkthroughs.

\section{Building Signals Datasets}
\label{sec:building-datasets}

As discussed in section \ref{subsec:signals-datasets}, \texttt{news-signals} provides an API for the creation of large-scale datasets representing collections of related signals. 

\paragraph{Bootstrapping Datasets using Wikidata} The Aylien NewsAPI links named entities in text to their Wikidata IDs \citep{wikidata-2014}. \texttt{news-signals} users can make use of the Wikidata Query Service\footnote{https://query.wikidata.org/} to easily build new datasets starting from SPARQL queries that return sets of matching entities \citep{2013sparql}. We build the datasets for NASDAQ-100, S\&P 500, and US Politicians in this manner, and the SPARQL queries used to bootstrap these entity sets are available in our repository.
For the purpose of this paper, and to exemplify use of the library, we build three example datasets: NASDAQ\-100, S\&P 500, and US Politicians. Each of these datasets is bootstrapped from a list of Wikidata entities belonging to the respective set. 
To retrieve the entity sets, we build a SPARQL query returning the set of Wikidata entities that match the query, and then use this entity set to generate a dataset. This is a powerful way to generate arbitrary datasets for collections of related entities: for example, datasets for all politicians from a particular country or all American football players could be generated in this fashion. 
Note that in some cases Wikidata does not contain all entities in a particular set, for example, the NASDAQ\-100 dataset contains fewer than 100 entities. Dataset statistics are summarized in Table \ref{table:news-signals-datasets-statistics}. Each of the entity sets is retrieved via one or more SPARQL queries\footnote{about SPARQL}. We then use the Aylien NewsAPI\footnote{https://aylien.com/} to sample up to $20$ stories about each entity for each day of the time period Jan 2020-Jan 2023. 

\paragraph{Multi-document Summarization (MDS)} 
We provide a multi-document summarization model in \texttt{news-signals} for turning clusters of news articles associated with a particular timestamp into an easily readable summary. In particular, we use a hybrid extractive-abstractive approach that first uses a centroid-based sentence extraction method \cite{ghalandari2017revisiting} to select 5 key sentences from the whole collection of provided news articles. We generate an abstractive summary from these sentences using a fine-tuned BART-large model \cite{lewis2020bart}. The model was fine-tuned on such extractive summaries on the WCEP dataset \cite{gholipour-2020-wcep}, which contains compact event summaries with a neutral style. 

\paragraph{Sampling News Data for Entities}
Importantly, we do not provide all news articles about each entity, rather, we provide only a sample of the news content about the entity for each day. This means that successful models should predict the time-series signal based upon the content of the article, rather than global numerical features \footnote{We may also consider models such as vector auto-regression that use signals derived from textual content as well as real-valued signals}. 

\paragraph{Connecting Entities with Timeseries Signals}

In our example datasets, we focus upon entities that exist in the Wikidata knowledge graph. Different time series signal sources can be automatically linked to these entities. The Wikimedia API itself exposes several interesting time series signals, such as the number of pageviews and the number of edits for each page. We hypothesize that these signals are affected by events occurring in the real world -- when an impactful event connected with an entity occurs, there is likely to be an observable change in signal behavior. 

\begin{table*}[ht]
\centering
\small
\begin{tabularx}{\textwidth}{|X|X|X|X|X|}
\hline
\textbf{Dataset Name} & \textbf{Start-End Date} & \textbf{Number of Signals} & \textbf{Total Articles} & \textbf{Time Series Targets} \\
\hline
US Politicians & 2020-01-01 to 2022-12-31 & 100 & 1285238 & news volume, Wikimedia pageviews \\
\hline
NASDAQ-100 & 2020-01-01 to 2022-12-31 & 99 & 1569139 & news volume, Wikimedia pageviews \\
\hline
S\&P-500 & 2020-01-01 to 2022-12-31 & 100 & 1728179 & news volume, Wikimedia pageviews \\
\hline
\end{tabularx}
\caption{Datasets Overview}
\label{table:news-signals-datasets-statistics}
\end{table*}

\subsection{Dataset Release}

To avoid potential licensing issues with releasing the news data content of the example datasets, at this stage we plan to only release the datasets containing article titles instead of full article texts and metadata. We also release a version of the datasets with daily abstractive summaries of the content, which do not reveal any source-specific content or data. Both versions will be available by email request to the authors\footnote{note also that all code used to produce the full datasets is open source}.

\paragraph{Extending NewsSignals}
Because our datasets are grounded on the Wikidata knowledge graph, they are easy to extend with new inputs, entities, and signals. Obvious extensions to our work might include textual data from platforms such as Twitter and Reddit, and market signals such as stock price or other technical indicators for entities that are connected with publicly traded companies. Datasets should also be easy to extend with additional entities, and we provide a set of tools for extending NewsSignals in the accompanying code repository\footnote{\url{https://github.com/AYLIEN/news-signals-datasets}}.

\subsubsection{Docker Container and Example K8s Configuration}

Because \texttt{news-signals} is designed to be used in both research and production settings, we have also provided a Dockerfile and an example Kubernetes (K8s) job configuration that can be deployed to Google Cloud Platform with minimal setup required. Together, these assets can be used to build signals datasets at a regular cadence, for example once a day or once a week.

\section{Example Models and Experiments}
\label{sec:experiements}

This section presents a suite of example models and experiments for users to quickly adapt to their own task settings, and to verify the utility of \texttt{news-signals} by establishing baselines for a straightforward anomaly prediction task.

\subsection{Binary Anomaly Prediction Task}
In this work, we focus on a simple binary anomaly prediction task, which we treat as text classification. The goal is to predict whether a time series signal about a particular entity is anomalous during some window in the past, present, or future, based on textual information in news feeds about the entity. The input for an individual prediction is a set of news articles, an \textit{aspect} (e.g. an entity) and the target a binary anomaly indicator. For simplicity, we predict the target value of a particular day from the textual input of the same day. 

We transform time series signals into binary anomaly predictions with the following procedure:

\subsection{Target Signals}
We experiment with two different time series target signals: anomalies time series of NewsAPI volume counts and Wikimedia page views. One target time series consists of day-level binary values for the time range of our datasets. We use a simple anomaly detector to convert the raw time series signals into binary values, based on the Z-score: We treat each value $x_t$ in a time series as an anomaly if the following is true: 
\begin{equation}
    \frac{x_t - \mu}{\sigma} > t
\end{equation}

\noindent where $\mu$ is the mean and $\sigma$ standard deviation of a time series. We set the anomaly threshold $t$ (measured in standard deviations) to 3 which results in a proportion of 1-3\% positive examples in our datasets. 

\subsection{Dataset Splits}

Each of the three dataset is split chronologically into training (80\%), validation (10\%) and test (10\%) sections. A trained model is informed about all entities in the training data and is tested to apply this knowledge to future data about these entities. The split can also be done across entities to test whether models can generalize to new entities. In this work, we focus on the simpler setting where the entities are known. Note that this does not apply to the zero-shot baselines using LLMs discussed below.

\subsection{Balanced Sampling for Training}
We preserve the validation and test dataset split as they are, i.e. with a small amount of 1-3\% of positive labels, and as continuous time periods. Since training with this label imbalance results in poor results, we create modified training datasets from the time period of the training split: we randomly sample 10,000 positive and 10,000 negative examples for each dataset.

\subsection{Compressing Textual Input}
Since we are dealing with a large amount of text for each individual prediction task, i.e. a set of 20 news articles, we need to compress these articles into a shorter text to fit the input size of typical current deep learning models. In our experiments, we use the concatenation of all headlines of a day as the textual input. We leave a comparison to alternatives, e.g. multi-document summaries or representative articles, to future work.

\subsection{Models for Anomaly Classification}
We include several text classification baselines that predict the target based on one day of compressed textual content:

\paragraph{Fine-tuned Transformer Classifier:} We fine-tune the pre-trained RoBERTa-base model \cite{liu2019roberta} with an un-trained randomly initialized binary classification head. We fine-tune the model on 1 epoch of the label-balanced training examples with a batch size of 8, a learning rate of 2e-5 and a weight decay of 0.01, using the Adam optimizer.

\paragraph{Random Forest with Sparse Lexical Features:}
We train random forest models on binary lexical features, to explore how well the target signals are represented in surface-level text. We use \texttt{sklearn}\footnote{\url{https://scikit-learn.org/}} to extract sparse binary token-indicator features, with a vocabulary of 10,000 tokens, excluding stop words. We train the models with 100 trees and a maximum depth of 20. We determined these values on the validation datasets.

\paragraph{Zero-Shot Classification with Llama-2 (13B):} We use Meta's \texttt{Llama-2-13b-chat}  \footnote{\url{https://huggingface.co/meta-llama/Llama-2-13b-chat}} model for zero-shot classification. We provide the 20 headlines of a day along with a prompt that describes the target signals as an input. The prompt used in the presented experiments is shown in Appendix \ref{sec:appendix-prompts}.

\subsection{Evaluation and Results}
We evaluate the binary anomaly classification task using Precision, Recall and F1-score. We put the results into perspective by comparing them to two random baselines: random-uniform, i.e. randomly classifying each input as an anomaly with a 50\% chance, and random-target, where we classify each input as an anomaly with a probability set to the proportion of positive examples in the test set. Table \ref{table:evaluation} shows the results for anomaly classification for news volume and Wikimedia pageviews as target signals. The trained models achieve above-random f1-scores on most of the dataset-target combinations, and obtain better results than the zero-shot baseline. We discuss the results in more detail in Appendix \ref{sec:appendix-results-discussion}. Figure \ref{fig:predicted-anomalies} shows an example of predicted anomalies, compared to the ground-truth anomalies defined by the anomaly detection method. The predicted anomalies in this example consistently correspond to a spike of Wikipedia page views on the day or shortly after the day on which the input news stories were published.

\begin{table*}[!ht]
\centering
\small
\begin{tabular}{|l|llllllllllll|}
\hline
\textbf{Target Signal} & \multicolumn{12}{c|}{\textbf{News Volume}}                                                                                                                                                                      \\ \hline
\textbf{Model/Dataset} & \multicolumn{4}{c|}{\textbf{Nasdaq-100}}                                   & \multicolumn{4}{c|}{\textbf{Smp-500}}                                      & \multicolumn{4}{c|}{\textbf{US-politicians}}          \\ \hline
                       & prec          & rec           & f1            & \multicolumn{1}{l|}{\%pos} & prec          & rec           & f1            & \multicolumn{1}{l|}{\%pos} & prec          & rec           & f1            & \%pos \\ \hline
Random - uniform       & 0.01          & 0.43          & 0.02          & \multicolumn{1}{l|}{0.5}   & 0.01          & 0.49          & 0.02          & \multicolumn{1}{l|}{0.49}  & 0.03          & 0.5           & 0.05          & 0.51  \\ \hline
Random - target        & 0.01          & 0.01          & 0.01          & \multicolumn{1}{l|}{0.01}  & 0.01          & 0.01          & 0.01          & \multicolumn{1}{l|}{0.01}  & 0.03          & 0.04          & 0.04          & 0.03  \\ \hline
Sparse + RF            & \textbf{0.19} & 0.58          & \textbf{0.28} & \multicolumn{1}{l|}{0.04}  & \textbf{0.12} & 0.30          & \textbf{0.18} & \multicolumn{1}{l|}{0.03}  & 0.20          & 0.28          & 0.23          & 0.04  \\ \hline
RoBERTa-base           & 0.12          & \textbf{0.69} & 0.2           & \multicolumn{1}{l|}{0.08}  & 0.1           & \textbf{0.52} & 0.17          & \multicolumn{1}{l|}{0.05}  & \textbf{0.21} & \textbf{0.69} & \textbf{0.33} & 0.08  \\ \hline
Llama-2-13b-chat       & 0.03          & 0.71          & 0.06          & \multicolumn{1}{l|}{0.16}  & 0.03          & 0.47          & 0.05          & \multicolumn{1}{l|}{0.1}   & 0.05          & 0.46          & 0.1           & 0.22  \\ \hline \hline
\textbf{Target Signal} & \multicolumn{12}{c|}{\textbf{Wikimedia Pageviews}}                                                                                                                                                              \\ \hline
Model/Dataset          & \multicolumn{4}{c|}{\textbf{Nasdaq-100}}                                   & \multicolumn{4}{c|}{\textbf{Smp-500}}                                      & \multicolumn{4}{c|}{\textbf{US-politicians}}          \\ \hline
                       & prec          & rec           & f1            & \multicolumn{1}{l|}{\%pos} & prec          & rec           & f1            & \multicolumn{1}{l|}{\%pos} & prec          & rec           & f1            & \%pos \\ \hline
Random - uniform       & 0.02          & 0.46          & 0.03          & \multicolumn{1}{l|}{0.5}   & 0.02          & 0.52          & 0.04          & \multicolumn{1}{l|}{0.49}  & 0.02          & 0.51          & 0.03          & 0.5   \\ \hline
Random - target        & 0.01          & 0.01          & 0.01          & \multicolumn{1}{l|}{0.02}  & 0.02          & 0.02          & 0.02          & \multicolumn{1}{l|}{0.02}  & 0.01          & 0.01          & 0.01          & 0.02  \\ \hline
Sparse + RF            & 0.01          & 0.09          & 0.02          & \multicolumn{1}{l|}{0.12}  & 0.02          & 0.11          & 0.04          & \multicolumn{1}{l|}{0.11}  & \textbf{0.22} & 0.16          & 0.19          & 0.01  \\ \hline
RoBERTa-base           & 0.01          & 0.09          & 0.03          & \multicolumn{1}{l|}{0.11}  & 0.03          & 0.19          & 0.05          & \multicolumn{1}{l|}{0.13}  & 0.18          & \textbf{0.57} & \textbf{0.28} & 0.05  \\ \hline
Llama-2-13b-chat       & \textbf{0.02} & \textbf{0.21} & \textbf{0.04} & \multicolumn{1}{l|}{0.3}   & \textbf{0.05} & \textbf{0.24} & \textbf{0.08} & \multicolumn{1}{l|}{0.24}  & 0.04          & 0.52          & 0.07          & 0.22  \\ \hline
\end{tabular}
\caption{Evaluation results for anomaly classification experiments. \%pos indicates the proportion of positive predicted labels.}
\label{table:evaluation}
\end{table*}

\begin{figure}[h]
\includegraphics[width=8cm]{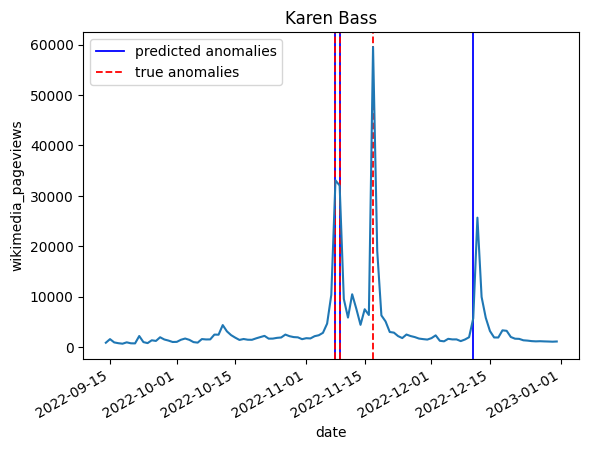}
\caption{Predicted and ground-truth anomalies of a Wikipedia pageviews time series of US politician Karen Bass. The predictions are from a random forest model with sparse lexical features.}
\label{fig:predicted-anomalies}
\end{figure}

\subsection{Extending to forecasting tasks}
This experimental setup can easily be converted into forecasting tasks by pairing the text content of a particular day with the target signal shifted by some offset into the future. By sliding our forecasting window earlier than the input, we can also study how well today's news predicts signals that already happened. This may be more relevant for signals that imply significant information asymmetry, such as stock price, as opposed to signals that are public by definition, such as Wikimedia pageviews. Rather than binary anomaly targets, we can train models to directly predict the real-valued signal or quantized representations of the signal.

\section{Intended Applications of NewsSignals}
\label{sec:news-signals-applications}

\paragraph{Time Series Forecasting using Texual Data}
As discussed, time series signal forecasting is an important task which is relatively unexplored in the context of models for natural language processing (NLP).

\paragraph{Financial Data Analysis} We believe that this dataset and task setting should be straightforward to adapt to financial time series analysis. Financial time series such as stock price and trading volume are impacted by real-world events. The behavior of market signals reflects sentiment about particular entities, and is influenced by events happening in the world. However, market signals may contain opaque and confounding factors that make accurate prediction more challenging. Although this work deliberately does not consider market signals, it is very straightforward to add market time series such as stock price(s) or trading volume to signals.

\paragraph{NLP for Healthcare}
The \texttt{text2signal} task setting is well-suited to the emerging field of BioNLP or NLP for Healthcare -- for example, predicting the number of hospital visits in subsequent months based upon a collection of doctor's notes from preceding months, or forecasting total medical expenditure in subsequent months based upon the content of a doctor's notes. 

\paragraph{Sentiment}

To date, sentiment analysis datasets have been created by human annotation. However, the annotation task is difficult to fully specify, and impossible to scale to real-world volumes of data. An insight is that there are many real world signals that can be considered proxies to sentiment, most obviously market signals, especially when the definition of sentiment is constrained to specific (entity, aspect) pairs. Instead of using model-derived sentiment to forecast time series, market signals can be used as ground-truth proxies to sentiment annotations. 

\paragraph{Social Sciences}
Social scientists may be interested in the tooling we have built around the Wikidata SPARQL endpoint, because \texttt{news-signals} allows users to easily build a set of signals connected to any set of Wikidata entities. In one of our example datasets, we produced a signal for every living US politician present in Wikidata, and we believe that many social scientists will be researching similar specific sets of entities and related time series signals.

This section discusses potential applications for \texttt{news-signals} and directions for future work.

\paragraph{Causality} News-signals may be useful for NLP researchers working on tasks related to causality, because time series signals are well-suited to causality research. In general, we wish to find out what types of information are likely to impact time series signals. Concretely, we may believe that there is a true causal relationship between news and the edit rate on Wikimedia pages.

\section{Related Work}
\label{sec:related-work}

\paragraph{NLP and Time Series Dataset Libraries}

\texttt{news-signals} can be seen as sitting between NLP-focused dataset libraries such as Huggingface Datasets \citep{lhoest-etal-2021-datasets} and time series focused libraries such as GluonTS and KATS \citep{gluon-ts,Jiang-KATS-2022}. We specifically build tooling for working with datasets with textual inputs and time series outputs, and \texttt{news-signals} is complementary to and compatible with other popular NLP and time series libraries.

\paragraph{Granger Causality}
It is natural to consider whether the content of textual inputs "caused" an observed time series signal behavior.
Granger causality \citep{granger-causality} is a method of measuring the degree to which one signal may cause another. 
\citet{marcinkevics2021interpretable} propose a framework for discovering Granger Causality with interpretable neural networks. 

Summary graphs \citep{PetersJanzingSchoelkopf17} are a useful way of compressing relationships about Granger causality.
\citet{Wen2017-mqrnn} introduce a flexible RNN architecture for time series forecasting. 
\citet{lm-framework-for-anomaly-detection} is a position paper discussing the use of PLMs for anomaly detection on financial data. 

\paragraph{Time Series prediction with Textual Inputs}
As discussed in Section \ref{sec:introduction}, one significant line of work focuses on predicting financial time series using signals derived from text, in particular aggregations of sentiment scores from social media posts \citep{finnlp-2021-financial,finnlp-2022-financial,arno-etal-2022-next,Li-2014-news-stocks,BingLi-twitter-sa-for-stock-price,kim2016-predicting-cryptocurrency,wang-luo-2021-predicting}, inter alia.

\paragraph{PLMs and Transfer Learning}

Recently, significant work has been done to adapt transformer-based models in particular to time series forecasting tasks with flexible semantics \citep{wen2023transformers}.

\paragraph{Timeline Summarization from News Corpora} A related line of work within the NLP community is constructing timelines of important events from large collections of news focused on long-term topics, e.g. disasters or entities \cite{martschat2018temporally}. The methods for identifying important events often make use of time-series-like signals defined over dates: the number of articles published per day or the number of times the date is mentioned in text \cite{binh2013predicting, ghalandari2020examining}.

\section{Conclusion}
\label{sec:conclusion}

We have presented \texttt{news-signals}, an open source library for building and working with NLP datasets that predict time series signals based on textual inputs. We hope that this library can be useful to a broad group of researchers and data scientists in both academic and industry settings. Naturally, we would be very happy for additional contributions from the open source community to further improve the library.

\bibliography{chris,anthology,custom}
\bibliographystyle{acl_natbib}




\appendix

\section{Discussion of Anomaly Classification Results}
\label{sec:appendix-results-discussion}

The trained models, i.e. RoBERTa-base and the random forest with sparse features achieve considerable improvements over random results on most of the dataset-target combinations, with mixed rankings. In these cases, the models detect 50-70 \% of the anomalies while only predicting 3-8\% anomalies in total, which is a promising pattern. All baselines show close-to-random results on Nasdaq-100 and Smp-500 with Wikimedia Pageviews. Zero-shot anomaly prediction with \texttt{Llama-2-13b-chat} generally performs worse than the trained models, but still better than the random baselines. Our zero-shot approach suffers from over-prediction of the positive class - a behavior that is difficult to tune when designing prompts. We leave more systematic prompt tuning for this task to future work.

\section{Prompting for Zero-Shot Approach}
\label{sec:appendix-prompts}

We use the following prompt template for \texttt{LLama-2-13b-chat} to do anomaly classification from news:

\textit{Headlines:
\{\{HEADLINES\}\}
The stories above all involve \{\{ENTITY\}\} and were published on the same day. Do these news stories indicate one of the most significant events for \{\{ENTITY\}\}? Respond with 'no' or 'yes'.}

We instantiate the placeholders with headlines and an entity name (person or company) for a specific data item.

We use the following system prompt: \textit{You are an anomaly detector for news.}

We formatted the prompt according to the \texttt{LLama-2}-specific pattern.

A key issue in this zero-shot approach is to control the overall proportion of times an anomaly is detected in a dataset, i.e. to express the significance or importance of news stories to entities. Signal-specific prompts, e.g. directly describing Wikipedia pageviews or news volume, turn out less effective than this generic description.

\end{document}